\documentclass{article}

\usepackage{arxiv}

\usepackage[utf8]{inputenc} 
\usepackage[T1]{fontenc}    
\usepackage{hyperref}       
\usepackage{url}            
\usepackage{booktabs}       
\usepackage{amsfonts}       
\usepackage{nicefrac}       
\usepackage{microtype}      
\usepackage{lipsum}
\usepackage{graphicx}
\graphicspath{ {./images/} }

\usepackage{algorithm}
\usepackage{algpseudocode}
\usepackage{graphics}
\usepackage{graphicx}

\usepackage{CJKutf8}
\usepackage{amsmath}
\usepackage{makecell}

\title{A Framework to Implement 1+N Multi-task Fine-tuning Pattern in LLMs Using The CGC-LoRA Algorithm}

\author{
 Chao Song \\
  OPPO Research Institute\\
  Shenzhen, Guangdong, China\\
  \texttt{songchao12@oppo.com} \\
   \And
 Zhihao Ye \\
  OPPO Research Institute\\
  Shenzhen, Guangdong, China\\
  \texttt{yezhihao3@oppo.com} \\
   \And
 Qiqiang Lin \\
  OPPO Research Institute\\
  Shenzhen, Guangdong, China\\
  \texttt{linqiqiang1@oppo.com} \\
  \And
 Qiuying Peng \\
  OPPO Research Institute\\
  Shenzhen, Guangdong, China\\
  \texttt{pengqiuying@oppo.com} \\
  \And
 Jun Wang \\
  OPPO Research Institute\\
  Shenzhen, Guangdong, China\\
  \texttt{wangjun7@oppo.com} \\
  \And
}

\begin{document}
\begin{CJK}{UTF8}{gbsn}
\maketitle
\begin{abstract}
With the productive evolution of large language models (LLMs) in the field of natural language processing (NLP), tons of effort has been made to effectively fine-tune common pre-trained LLMs to fulfill a variety of tasks in one or multiple specific domain (e.g., code interpreting, data analysis, medicine, and item recommendation). In practice, there are two prevailing ways, in which the adaptation can be achieved: (i) \textit{Multiple Independent Models:} Pre-trained LLMs are fine-tuned a few times independently using the corresponding training samples from each task. (ii) \textit{An Integrated Model:} Samples from all tasks are employed to fine-tune a pre-trianed LLM unitedly. The first scheme, in most cases, leads to high computing cost since pre-trained LLMs need to undergo fine-tuning process numerous times separately. Although various parameter efficient fine-tuning (PEFT) methods are proposed and they indeed can alleviate the issue in a sense, high computing cost is still prohibitively and unacceptable. More seriously, the knowledge of each task stays isolated and it cannot be shared across different tasks efficiently, which may lead to the sub-optimal overall performance of fine-tuned LLMs. Regarding the second method, although different kinds of loss functions are ingeniously designed, which sufficiently consider especial situations of multi-task learning (MTL), a promotive consequence of one task frequently causes the degradation of other one or ones, called the seesawing issue. To address the high computing cost and seesawing issue simultaneously, we propose a unified framework that implements a $1+N$ mutli-task fine-tuning pattern in LLMs using a novel Customized Gate Control (CGC) Low-rank Adaptation (LoRA) algorithm. In detail, ``1" stands for a central LLM offering common and extensive capabilities in general cases while ``$N$" denotes N sets of CGC-LoRA modules that provides adaptive abilities to behave well on multiple clusters of unseen tasks, respectively. Our work aims to take an advantage of both MTL (i.e., CGC) and PEFT (i.e., LoRA) scheme. For a given cluster of tasks, we design an innovative layer that contains two types of experts as additional trainable parameters to make LoRA be compatible with MTL. In specific, task-common experts intend to capture the message that should be shared across distinct tasks while service objects of task-specific experts are specific tasks. In addition, we also nominate a Task-motivated Gate (TMG) function that determines contributions of each expert to different tasks and this function desires to organize experts in an efficient way. To comprehensively evaluate the proposed framework, we conduct well-designed experiments on two public datasets including a wide range of multi-tasks: (i) Prompt Chinese Biomedical Language Understanding Evaluation (PromptCBLUE). (ii) Firefly. Both datasets involve over a thousand samples of each sub-task. The experimental results demonstrate that the unified framework with CGC-LoRA modules achieves higher evaluation scores than all benchmarks on both two datasets.
\end{abstract}


\section{Introduction}
Large language models (LLMs) have made great progress recent two years \cite{Zhao2023, Chang2023, Naveed2023} and numerous versions of LLMs, namely, ChatGPT \cite{Liu2023, Long2022training}, LlaMa \cite{Touvron2023}, ERNIE \cite{Sun2021}, and ChatGLM \cite{Zeng2022}, sprung up. Worldwide high-tech companies and non-profit research institutes are paying significant attention to the evolution of LLMs since they have already shown typically impressive performance on varied applications. For example, Wang and Thai in \cite{Wang2023, Thai2022} demonstrate that LLMs get outstanding achievements on document-level machine translation and meanwhile LLMs is proved by \cite{Deng2023, Lai2023} to be equipped with dramatic multilingual learning capability. Besides, LLMs present remarkable talent on multi-modal operation, such as precisely producing websites from handwritten text and determining humorous components within images \cite{Zhu2023}. As for recommendation system, LLMs also indicates a non-negligible potential on a variety of tasks, including sequential recommendation, rating prediction, explanation generation, review summary, direct recommendation \cite{Geng2022, Wu2023, Liu2023is}. Lastly but even more incredibly, Wei and Huang in \cite{Wei2022, Huang2022} announce that the reasoning ability of LLMs can even be elicited by a chain-of-thought (CoT) algorithm.

The majority of LLMs, especially for those extraodinarily authoritative ones like ChatGPT and LlaMa, are pre-trained on general purposes using universal data \cite{Long2022training, Liu2023, Touvron2023}. Although the few-shot learning and even zero-shot learning capability of LLMs can be stimulated via in-context learning \cite{Agrawal2022, Hegselmann2023, Izacard2022}, the adaptation to a specific territory by fine-tuning is consistently favourable \cite{Wang2023huatuo, Wang2023codet5}. For instance, Wang in \cite{Wang2023huatuo} determines that fine-tuned large language model, named after Hua Tuo, outperforms the original LlaMa on Chinese medical knowledge. Additionally, CodeT5+ \cite{Wang2023codet5}, a follow-on version of T5 fine-tuned on coding-related data, can achieve new SoTA results on the HumanEval, a specialized dataset on coding-related tasks \cite{Chen2021evaluating}, against other fundamental pre-trained LLMs including the basic version of T5 model. Therefore, in this work, we engage in the open-source LLMs and fine-tune them with task-specific datasets.

To a great extent, two inevitable dilemmas will occur when applying fine-tuning process to pre-trained LLMs. Firstly, a certain field, in most situations, involves a large numbers of heterogeneous tasks and consequently, a general fine-tuned LLMs cannot cover all tasks completely, called \textit{Various Task Impact} \cite{Chang2023, Liu2023moelora}. Considering the real-world advertisement recommendation area, it encompasses a wide range of assignments, like sequential recommendation, rating prediction, explanation generation, review summary, direct recommendation, etc., and these diverse sub-tasks usually have objectives of similarity at different levels and even of conflict \cite{Geng2022, Liu2023is}. To overcome this challenge, two feasible solutions are presented: (i) \textit{Multiple Independent Models:} Fine-tuning an independent model for each particular task is practical, however, this strategy appeals considerable professional knowledge and a mass of labor. Also, information is precisely isolated and they cannot be shared across each task smoothly \cite{Ma2018modeling, Tang2020progressive}. What is more important, full parameters of a few fine-tuned LLMs for different tasks need to be saved that is unreasonable and even absurd, especially considering applications on mobile device due to the memory limitation. (ii) \textit{An Integrated Model:} Fine-tuning an integrated model using samples from all tasks simultaneously can solve the information isolation and memory limitation problems mentioned above. As for this strategy, although knowledge is not segregated, at least in theory, the seesawing problem is frequently reported and it signifies that the enhancement of one task leads to the degradation of others \cite{Ma2018modeling, Tang2020progressive}. Secondly, \textit{High Computing Cost} is another challenge. Since the LLMs having billions of parameters dominate nearly all areas, fine-tuning full parameters becomes extremely costly and utopian \cite{Long2022training, Sun2021}. Even if any companies or institutes can afford such astronomical computing cost, a limited amount of available data will ultimately cause over-fitting \cite{Mishra2022}. As a result, to have LLMs work excellently on applications of a wide variety, these two problems are urgently to be solved preferably.

Considering the \textit{Various Task Impact}, different kinds of multi-task learning (MTL) models have been proposed \cite{Zhang2018an, Thung2018}, among which hard-parameter-sharing structure is designed primarily \cite{Caruana1997}. Although it, to some degree, promotes knowledge circulation among each task, negative transfer problem is exposed as parameters are straightforwardly and even wildly shared. To solve this issue, cross-stitch network \cite{Misra2016} and sluice network \cite{Ruder122017} are announced by Misra and Ruder12, separately. Both networks can merge information from various tasks selectively as the combination is controlled linearly by a set of weights. Nonetheless, since weights stay the same for all samples, the seesawing phenomenon cannot be completely addressed. Next, the gate structure and both cross- and self-attention mechanisms are demonstrated to further solved the seesawing issue. Additionally, mixture-of-expert (MOE) related models \cite{Jacobs1991, Ma2018modeling} first proposes to combine task-common experts by fusing the gate module and the attention module. In particular, both expert and attention modules are shared among all tasks while no task-specific module is scheduled. In contrast, Tang in \cite{Tang2020progressive} releases a customized gate control (CGC) module, which separates task-common and task-specific experts explicitly to keep clear of parameter disagreement caused by complicated correlation among tasks. Such an explicit design in virtue of professional prior knowledge is conducive to determine the convergence direction more accurately in practice \cite{Liu2019end}. Besides the \textit{Various Task Impact} problem, the \textit{High Computing Cost} is another roadblock when MTL frameworks are applied to fine-tune LLMs since current frameworks are principally appropriate to full-parameter fine-tuning algorithms. Fortunately, a class of fine-tuning methods called parameter efficient fine-tuning (PEFT), which freeze parameters of pre-trained LLMs and instead only tune a delimited quantity of additional parameters, are explored to solve the high computing cost issue \cite{Hu2021lora, Liu2022ptuning}. Among proposed PEFT approaches, low-rank adaptation (LoRA) is a typical one and it suggests to just train a set of paired low-rank matrices to provide LLMs with adaptation capacity\cite{Hu2021lora}. However, making PEFT methods be compatible with MTL frameworks is still an open question \cite{Liu2023moelora}.

To solve the issues of \textit{Various Task Impact} and \textit{High Computing Cost} simultaneously by integrating MTL and PEFT, we proposed a universal framework to implement $1+N$ multi-task fine-tuning pattern in LLMs. Specifically, a large range of tasks are divided into multiple clusters based on the professional prior knowledge or Inter-Task Affinity proposed by \cite{Fifty2021efficiently} and for a certain cluster, a central LLM is fine-tuned to adjust to these tasks by LoRA of a multi-task edition. This central LLM with sets of plug-in LoRA modules serves as the ultimate model that can handle diverse tasks from different fields with ease. As an essential component of the whole framework, we further propose an innovative unified LoRA and CGC structure, name after CGC-LoRA, inspired by MOE-LoRA \cite{Liu2023moelora} and it provides the framework with an effective and efficient fine-tuning performance in MTL situations. In detail, thanks to the explicit separation of task-common and task-specific experts, CGC is proved to have a practically preferred convergence behavior \cite{Tang2020progressive} and we demonstrate that such a valuable character can be smoothly transferred to the CGC-LoRA architecture. As the supplement to the basic edition of LoRA where a singular set of paired low-rank matrices are fine-tuned, CGC-LoRA incorporates multiple sets of paired low-rank matrices that stand for task-common and task-shared experts. Moreover, to keep the counts of trainable parameters the same, the summation of rank of task-common and task-specific matrices is equal to the rank of the original LoRA scheme. Furthermore, to advance the fusion capacity of experts, a task motivated gate function is proposed to adaptively determine the contribution of those two types of experts to each task only based on the task ID of the input sample. Eventually, the contributions of this work are summarized as follows:
\begin{itemize}
\item We announce a framework to implement $1+N$ multi-task fine-tuning pattern in LLMs, which can equip general pre-trained LLMs with adaptation to a large range of unseen tasks. It is worthy noting that this is a universal framework and it can be compatible with LoRA of any multi-task editions (e.g., MOE-LoRA \cite{Liu2023moelora}, LoRAHub \cite{Huang2023lorahub}, and proposed CGC-LoRA).
\item We associate the power of both CGC structure and LoRA scheme by designing an ingenious unified MTL PEFT architecture. Besides, to lubricate the cooperation between task-common and task-specific experts, a task motivated gate function is proposed to regulate the weights of all experts and completely, determine the distinct contributions of experts to each task.
\item We conduct comprehensive experiments on two public multi-task datasets: (i) the Prompt Chinese Biomedical Language Understanding Evaluation (PromptCBLUE) dataset \cite{Zan2021building, Zhu2023promptcblue}; (ii) the Firefly dataset \cite{Firefly}. The experimental results demonstrate the superiority of the proposed framework over other long-tested baselines. To the best of our knowledge, this paper represents the first effort to evaluate the dominance of the framework equipped with a MTL-PEFT-unified structure for LLM-driven models on vast of applications using multiple public datasets.
\end{itemize}

\section{Preliminary}
In this section, a compact introduction to how LLMs can be applied to different kinds of applications in various domains is represented. First, an overview of the proposed framework that is to implement $1+N$ multi-task fine-tuning pattern in LLM is offered in Section \ref{framework overview} and next, we describe the details of input pre-processing along with those of output post-processing in Section \ref{input}. 

\begin{figure*}
  \includegraphics[scale=0.45]{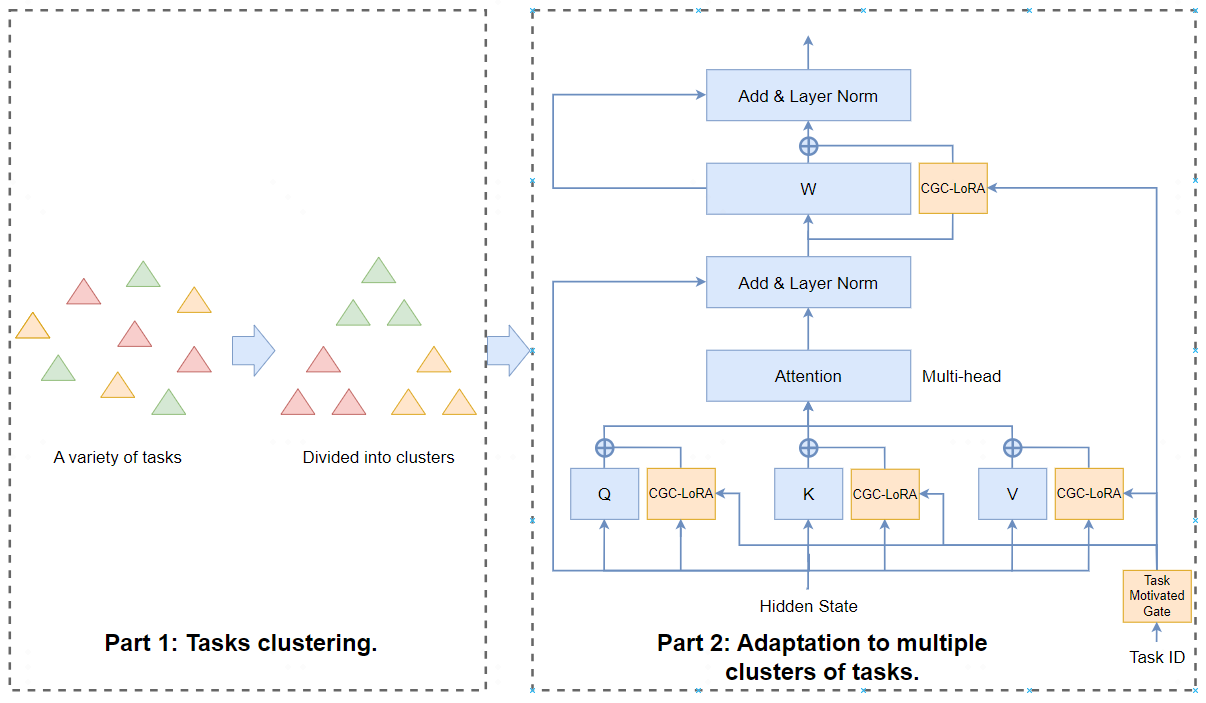}
  \caption{An overview of the proposed framework to implement $1+N$ multi-task fine-tuning pattern in LLMs. The first step is to divide a variety of tasks into $N$ clusters based on professional prior information or Inter-Task Affinity \cite{Fifty2021efficiently} and the second step is to adapt a central LLM to each cluster of tasks using the LoRA fine-tuning algorithm of multi-task editions (e.g., MOE-LoRA, LoRAHub, and CGC-LoRA).}
  \label{fig:framework}
\end{figure*}

\subsection{An Overview of the Framework} \label{framework overview}
In this section, we present an overview of the proposed framework. It introduces how the framework implements $1+N$ multi-task fine-tuning pattern in LLMs to adapt a central LLM to clusters of tasks using the LoRA fine-tuning algorithm of multi-task editions (e.g., MOE-LoRA, LoRAHub, and CGC-LoRA). The entire framework can be divided into two primary parts: (i) \textit{Tasks clustering:} As depicted in Figure \ref{fig:framework}, a LLM-driven model may face with a variety of tasks from different domains (e.g., code interpreting, data analysis, medicine, and item recommendation). Among them, tasks in certain fields highly have strong similarity, like code interpreting and data analysis while tasks in other fields, such as data analysis and item recommendation, almost have no relations and even conflicts. According to the assumptions in MTL, information sharing among related tasks can, to some degree, relieve the negative effect from lack of data and also benefit the effectiveness and efficiency of model training process and vice versa. As a consequence, tasks are grouped into $N$ clusters at the first stage. (ii) \textit{Adaptation to multiple clusters of tasks:} Following tasks clustering, the LoRA fine-tuning algorithm of multi-task editions is applied to each cluster and finally a central LLM with $N$ sets of multi-task LoRA modules serves as an integrated model that can offer service to various tasks from all domains by switching to the corresponding module. Since the fine-tuning process for each cluster is repetitive,  we set $N$ to be 1 in the remaining part for convenience and also without loss of generality.

\begin{figure*}
  \centering
  \includegraphics[scale=0.7]{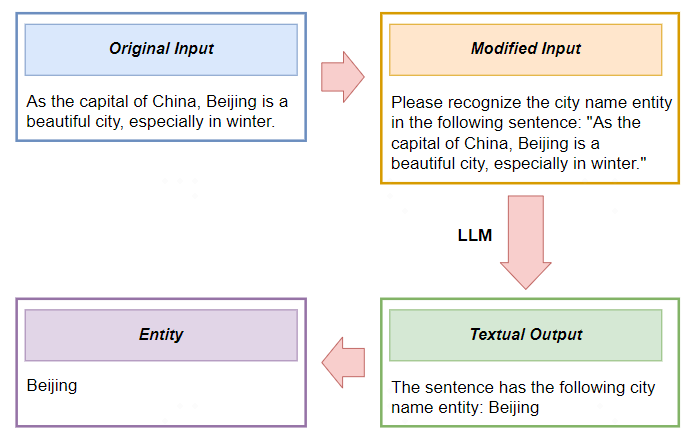}
  \caption{A name entity recognition (NER) example to illustrate the whole pipeline of applications by large language models (LLMs). It includes the pre-processing to a conventional input, the textual answer of a LLM, the post-processing to extract final results.}
  \label{fig:input2output}
\end{figure*}

\subsection{Input and Output Modification} \label{input}
Except for multi-modal LLMs, the input and output of LLMs should be in text pattern \cite{Geng2022, Zan2021building, Zhu2023promptcblue}, which leads to a typically different prototype from traditional models. As a consequence, there is a prerequisite to customize the input/output format to be compatible with LLMs. As shown in Figure \ref{fig:input2output}, Name Entity Recognition (NER) task \cite{Yadav2019a}, as a concrete example, suggests the details of the reformulation of input and output.

\textbf{Input Modification}. Besides the initial textual input, auxiliary instruction templates that can guide LLMs in accomplishing the given task should be attached. In particular, as characterized in Figure \ref{fig:input2output}, the input is formatted as: \textit{``Please recognize the name entity in the following sentence: [Text]''}, represented by $I_{modified}$, where \textit{[Text]} acts as a placeholder for the original input, denoted as $I_{original}$.

\textbf{Output Modification}. Outputs of traditional models in natural language processing (NLP) are commonly at word-level \cite{Devlin2018bert, Peng2019transfer} and as for LLMs, we integrate target entities $\{e_{1},...,e_{N_{e}}\}$ into linguistic texts. As shown in Figure \ref{fig:input2output}, the output of NER task can be formatted as following: ``\textit{The text has the following entities: $e_{1},...,e_{N_{e}}$''}, represented by $O_{sentence}$.

In summary, the whole paradigm can be described as follows:

\begin{equation} \label{preprocess}
  I_{original}\rightarrow I_{modified}\stackrel{LLM}{\longrightarrow} O_{sentence}\rightarrow e_{1},...,e_{N_{e}}
\end{equation}

With the transformation of input and output, the pre-trained LLMs, such as LlaMA \cite{Touvron2023}, ChatGLM \cite{Zeng2022}, etc., can be straightforwardly fed by purely lingual data. After fine-tuning, LLMs can routinely generate answers by following the pipeline formulated as Equation (\ref{preprocess}).

\section{Problem Formulation}
As described in Section \ref{framework overview}, sorts of territories often envelop a variety of tasks in practice, for example, (i) Item Recommendation: Sequential recommendation, rating prediction, explanation generation, etc. (ii) Medical Application: Name entity recognition, attribute extraction, report generation, etc. (iii) Text Generation: Lyric generation, composition generation, poetry generation, etc. The purpose of this work is to fine-tune a pre-trained central LLM to meet the requirements from multiple domains, each of which contains numerous sub-tasks. Considering that the numbers of clusters has been set to $N=1$ for convenience, a cluster of tasks are denoted as $\mathbb{T} = \{ \mathcal{T}_{1},...,\mathcal{T}_{i},...,\mathcal{T}_{N_{T}} \}$ while the input and output data of $\mathcal{T}_{i}$ can be described as $\mathcal{I}_{i} = \{ I^{\mathcal{T}_{i}}_{k} \}^{N_{\mathcal{T}_{i}}}_{k=1}$ and $\mathcal{O}_{i} = \{ O^{\mathcal{T}_{i}}_{k} \}^{N_{\mathcal{T}_{i}}}_{k=1}$, respectively. In specific, $N_{\mathcal{T}_{i}}$ stands for the numbers of paired data (i.e., linguistic input and output data) of $\mathcal{T}_{i}$. To be brief, the superscript $\mathcal{T}_{i}$ will be left out in the following essay. In detailed, the problem of multi-task fine-tuning can be formulated as: Given the paired input and output data of all tasks $\{ I_{i}, O_{i} \}^{N_{T}}_{i=1}$, the target is to optimize the parameters $\Theta$ of a pre-trained LLM to achieve the shining behavior among all tasks of distinguished similarity and it is noteworthy that $\Theta$ alternately stands for full parameters or selective ones. 

Regarding the objective when fine-tuning the pre-trained LLMs, we straightforwardly apply a common one in the field of natural language processing (i.e., the conditional language modeling objective) after normalizing input and output data of all tasks into a steady linguistic format as detailed in Section \ref{input}. Besides, in order to stimulate the information sharing across each task, all training samples are fed into the LLMs simultaneously and as a consequence, the objective function of LLMs fine-tuning in a multi-task case can be written as:

\begin{equation} \label{loss}
  \max_{\Theta} \sum_{i \in [N_{T}]} \sum_{x \in \mathcal{I}_{i}, y \in \mathcal{O}_{i}} \sum^{|y|}_{t=1}log(P_{\Theta}(y_{t}|x,y \leq t))
\end{equation}

\section{CGC-LoRA}
In this section, a comprehensive illustration of CGC-LoRA structure is represented. In Section \ref{overview}, a description over the whole fine-tuning process is primarily conducted and as follows, Section \ref{cgc} characterizes how CGC-LoRA takes advantage of both CGC and LoRA scheme in detail. Subsequently, a prominent component of CGC-LoRA network (i.e., a task-motivated gate function), which controls the contribution of both task-common and task-specific experts to each task is illustrated in Section \ref{task}. Finally, an elaborate definition of training and inference pipeline is given in Section \ref{training}.

\subsection{Overview} \label{overview}
\begin{figure}
  \centering
  \includegraphics[scale=0.5]{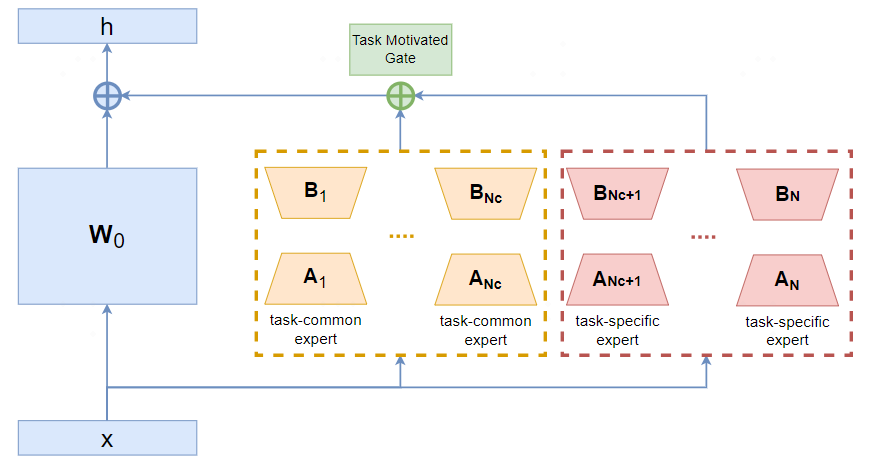}
  \caption{The detailed architecture of the proposed CGC-LoRA scheme. Specifically, the parameters of modules in blue are frozen while those of components in green, yellow, red are trainable during the fine-tuning process.}
  \label{fig:cgc}
\end{figure}

As mentioned, parameter efficient fine-tuning (PEFT) algorithm provides an innovative way, in which LLMs can be fine-tuned productively and efficiently. More concretely, only parameters of additional paired low-rank matrices are trainable while primordial parameters of pre-trained LLMs stay frozen. The extra parameters offer updated knowledge to dense layers. Inspired by MOE-LoRA \cite{Liu2023moelora}, our approach blends CGC network with LoRA strategy and the combination is attached to \textit{Query module}, \textit{Key module}, \textit{Value module}, and \textit{dense layers} in the Transformer architecture as displayed in Figure \ref{fig:cgc}. Furthermore, the proposed CGC-LoRA structure not only inherits the main advantage of PEFT algorithm (i.e., productiveness and efficiency) but also extends to solve the task variety problem and seesawing issue in multi-task learning (MTL) cases. Specifically, each CGC-LoRA layer incorporates two types of experts: (i) Task-common experts: They aim to capture conjunct knowledge across various tasks. (ii) Task-specific experts: Their responsibility is to extract specific information of each task. More details is depicted and elaborated in Figure \ref{fig:cgc} and Section \ref{cgc}, separately. Additionally, as an unimpressive but significant component of CGC-LoRA layer, a task-motivated gate module determines the contribution weights of two types of experts (i.e., task-common experts and task-specific experts) and more precisely, the outcome of each task-motivated gate only depends on task ID of each task rather than any other features of input samples. In Section \ref{task}, more details of how task-motivated gate modules are integrated into a CGC-LoRA layer can be found. Lastly, another implementation detail is worth noting that gate functions of all CGC-LoRA layers can be shared or independent, which, in a sense, influences the fitting ability of fine-tuned LLMs.

\subsection{CGC-LoRA} \label{cgc}
Thanks to the characteristic of both efficiency and effectiveness, Low-rank Adaptation (LoRA) gradually becomes a mainstream and prominent fine-tuning algorithm on Transformer-based networks \cite{Hu2021lora}. Since the efficiency of our framework comes from the essential theorem of LoRA, we first briefly review its principle that can make the comprehension of the proposed CGC-LoRA be simpler.

As for LoRA, a low-rank decomposition is applied to the parameters fine-tuning procedure in LLMs and more precisely, full-rank matrices that traditionally stands for full parameters are decomposed into paired low-rank matrices as the equation $\mathbf{W}_{0} + \Delta \mathbf{W} = \mathbf{W}_{0} + \mathbf{BA}$. It is worthy noting that $\mathbf{W}_{0} \in \mathbb{R}^{d_{in} \times d_{out}}$ denotes the parameter matrices of a pre-trained central LLMs while $\Delta \mathbf{W} \in \mathbb{R}^{d_{in} \times d_{out}}$ represents the additional matrices, which offers adaptation capability to pre-trained LLMs. In addition, $\mathbf{B} \in \mathbb{R}^{d_{in} \times r}$ and $\mathbf{A} \in \mathbb{R}^{r \times d_{in}}$ stands for a pair of decomposed trainable low-rank matrices. Built upon these notations, an abstract expression of a dense layer paired with the LoRA scheme can be formulated as:

\begin{equation} \label{lora}
\begin{split}
  \mathbf{h} &= \mathbf{W}_{0} \mathbf{x} + \frac{\alpha}{r} \cdot \Delta \mathbf{W} \mathbf{x} \\
  &= \mathbf{W}_{0} \mathbf{x} + \frac{\alpha}{r} \cdot \mathbf{B} \mathbf{A} \mathbf{x}
\end{split}
\end{equation}

where $\mathbf{x}$ and $\mathbf{h}$ stand for the input and output vector of dimension $d_{in}$ and $d_{out}$, respectively. Moreover, the numbers of additional trainable parameters is determined by $r$, the rank of decomposed low-rank matrices while $\alpha$ is designed to alleviate the impact caused by the variance of rank $r$. As emphasised, all parameters (e.g., $\mathbf{W}_{q}, \mathbf{W}_{k}, \mathbf{W}_{v}$ and $\mathbf{W}$) of pre-trained LLMs stay frozen during the fine-tuning process as illustrated in Figure \ref{fig:cgc}. Instead, the paired low-rank matrices, $\mathbf{A}$ and $\mathbf{B}$, are trained from scratch. As suggested by \cite{Hu2021lora}, $\mathbf{A}$ is initialized by a random Gaussian while $\mathbf{B}$ is set to be zero. Considering that $r \ll d_{in}$ and $r \ll d_{out}$, the total numbers of trainable parameters in $\mathbf{A}$ and $\mathbf{B}$ is typically fewer than that in $\mathbf{W}_{0}$, which is the principle mechanism stimulates the efficiency and effectiveness during the fine-tuning process.

Although PEFT algorithms (i.e., LoRA) can solve the high computing cost issue because of the demand for full-parameter fine-tuning excellently, they are at a loss how to achieve a efficient knowledge circulation in MTL cases. As for LoRA, additional parameters are fine-tuned as a whole for all tasks by feeding all samples into the pre-trained LLMs simultaneously. To further optimize the fine-tuning algorithm, an ingenious plan, enlightened by the multi-expert structure, is to partition matrices $\mathbf{A}$ and $\mathbf{B}$ of rank $r$ into several segments $\{ \mathbf{A}_1,...,\mathbf{A}_{N} \}$ and $\{ \mathbf{B}_1,...,\mathbf{B}_{N} \}$, rank summation of which equals to $r$. In this work, we propose a innovative network layer, named by CGC-LoRA, which uses a Customized Gate Control (CGC) structure as reference. Particularly, CGC employs two types of expert networks (i.e., task-common and task-specific experts) to apprehend both shared and isolated knowledge of multi-task. Such a pattern can be integrated into a LoRA layer by a decomposition of matrices $\mathbf{A}$ and $\mathbf{B}$ and naturally, segments $\{ \mathbf{A}_1,...,\mathbf{A}_{N} \}$ and $\{ \mathbf{B}_1,...,\mathbf{B}_{N} \}$ corresponds to each task-common and task-specific expert, denoted as $E^{C} \in \{ E^{C}_{i} \}^{N_{C}}_{i=1}$ and $E^{S} \in \{ E^{S}_{i} \}^{N_{S}}_{i=1}$, respectively. Here $N_{C}$ and $N_{S}$ stand for the numbers of task-common and task-specific expert, separately and the summation of $N_{C}$ and $N_{S}$ is equal to $N$, the total numbers of experts. Generally speaking, the numbers of task-specific experts is the same as that of tasks (i.e., $N_{C} = N_{T}$). During the CGC-LoRA fine-tuning process, additional parameters, serving as experts, are trained using data from all tasks and correspondingly, task-common and task-specific experts intrinsically capture shared and specific information. Given this architecture, a forward process of a CGC-LoRA layer for samples from task $\mathcal{T}_{j}$ can be formulated as:

\begin{equation} \label{cgc_forward}
\begin{split}
  \mathbf{h}_{j} &= \mathbf{W}_{0} \mathbf{x}_{j} + \frac{\alpha}{r} \cdot \Delta \mathbf{W}_{j} \mathbf{x}_{j} \\
  &= \mathbf{W}_{0} \mathbf{x}_{j} + \frac{\alpha}{r} \cdot [w^{S}_{j} \cdot E^{S}_{j}(\mathbf{x}_{j}) + \sum^{N_{C}}_{i=1} w^{C}_{ji} \cdot E^{C}_{i}(\mathbf{x}_{j})] \\
  &= \mathbf{W}_{0} \mathbf{x}_{j} + \frac{\alpha}{r} \cdot [w^{S}_{j} \cdot \mathbf{B}^{S}_{j} \mathbf{A}^{S}_{j} + \sum^{N_{C}}_{i=1} w^{C}_{ji} \cdot \mathbf{B}^{C}_{i} \mathbf{A}^{C}_{i}] \mathbf{x}_{j}
\end{split}
\end{equation}

where $\mathbf{h}_{j}$ and $\mathbf{x}_{j}$ stands for the input and output of intermediate LLM layers (e.g., a dense layer in a Transformer structure) for samples from $\mathcal{T}_{j}$, respectively. A pair of $\mathbf{B}^{S}_{j} \in \mathbb{R}^{d_{in} \times \frac{r}{N}}$ and $\mathbf{A}^{S}_{j} \in \mathbb{R}^{\frac{r}{N} \times d_{out}}$ represents a task-specific expert for task $\mathcal{T}_{j}$ while $N_{C}$ pairs of $\mathbf{B}^{C}_{i} \in \mathbb{R}^{d_{in} \times \frac{r}{N}}$ and $\mathbf{A}^{C}_{i} \in \mathbb{R}^{\frac{r}{N} \times d_{out}}$ denotes the task-common experts. For convenience, in Equation (\ref{cgc_forward}) the rank of $\mathbf{A}^{S} \in \{ A^{S} \}^{N_{S}}_{j}$, $\mathbf{B}^{S} \in \{ B^{S} \}^{N_{S}}_{j}$, $\mathbf{A}^{C} \in \{ A^{C} \}^{N_{C}}_{i}$, and $\mathbf{B}^{C} \in \{ B^{C} \}^{N_{C}}_{i}$ are all set to be $\frac{r}{N}$ equally. In a more general expression, the rank of $A^{S}_{j}$, $B^{S}_{j}$, $A^{C}_{i}$, and $B^{C}_{i}$ can be different as long as $rank(A^{S}_{j}) = rank(B^{S}_{j})$ and $rank(A^{C}_{i}) = rank(B^{C}_{i})$ and also $\sum_{j=1}^{N_{S}}rank(A^{S}_{j}) + \sum_{i=1}^{N_{C}}rank(A^{C}_{i}) = r$. Furthermore, the contribution weight of each expert are task-specific and in specific, weights are determined by our proposed task-motivated gate function, which will be specified in the following section.

In this paragraph, we will compare the numbers of trainable parameters of basic LoRA and proposed CGC-LoRA. As for LoRA, trainable parameters are fully included in paired low-rank matrices  $\mathbf{B} \in \mathbb{R}^{d_{in} \times r}$ and $\mathbf{A} \in \mathbb{R}^{r \times d_{out}}$ and consequently, the total counts of trainable parameters in LoRA can be calculated as $d_{in} \times r + r \times d_{out} = r \times (d_{in} + d_{out})$. Comparably, suppose the numbers of task-common and task-specific experts equals to $N_{C}$ and $N_{S}$, respectively. Totally, the trainable parameters of all experts can be expressed as $N_{C} \times \frac{r}{N} \times (d_{in} + d_{out}) + N_{S} \times \frac{r}{N} \times (d_{in} + d_{out}) =  \frac{N_{C} + N_{S}}{N} \times r \times (d_{in} + d_{out}) = r \times (d_{in} + d_{out})$ considering $N_{C} + N_{S} = N$. As emphasized above, the equation illustrates that the limit of rank $r$ is divided into each expert evenly while the conclusion that CGC-LoRA has the same counts of trainable parameters as LoRA holds constant even if the division is uneven.

\subsection{Task-motivated Gate Function} \label{task}
In this section, we elaborate how weights of those two types of experts are determined by the proposed task-motivated gate function. As previously mentioned, the information learned by task-common experts should be shared across each task while task-specific experts need to focus on a given task. In practice, the contributions of those two kinds of experts to a certain task is determined by weights $w^{S}_{j}$ and $w^{C}_{ji}$, which correspond to task-specific and task-common experts, separately. To achieve this, a task embedding matrix, denoted as $\mathbf{E} \in \mathbb{R}^{|\mathbb{T}| \times d_{T}}$, aims to represent task IDs in a hyper-dimensional space and here $d_{T}$ stands for the dimension of the task embedding. For a certain task $\mathcal{T}_{j}$, the \textit{j}-th column of $\mathbf{E}$ is extracted, which serves as the embedding vector for that task, denoted as $\mathbf{e}_{j} \in \mathbb{R}^{d_{T}}$. As for the weights of task-common experts (i.e., $\mathbf{w}^{C}_{j} = [w^{C}_{j1},...,w^{C}_{ji},...,w^{C}_{jN_{C}}]^{T}$), the formulation can be expressed as follows:

\begin{equation} \label{wcj}
   \mathbf{w}^{C}_{j} = \mathbf{W}^{C}_{T}\mathbf{e}_{j}
\end{equation}

where $\mathbf{W}^{C}_{T} \in \mathbb{R}^{N_{C} \times d_{T}}$ stands for a transformation matrix that can complete a linear transformation of task embedding of a certain task-common expert. Similarly, considering the weights of task-specific experts, since each expert corresponds to its own task and has none contribution to the other tasks, the weight $w^{S}_{j}$ can be expressed as follows:

\begin{equation} \label{wsj}
   w^{S}_{j} = \mathbf{W}^{S}_{T}\mathbf{e}_{j}
\end{equation}

where $\mathbf{W}^{S}_{T} \in \mathbb{R}^{1 \times d_{T}}$ denotes another transformation vector that applies a linear transformation of task embedding of a given task-specific expert. Finally, we apply a $\mathrm{Softmax}$ operation after the linear transformation to normalize the weights of both task-common and task-specific experts as follows:

\begin{equation} \label{softmax}
   \mathbf{w}_{j} = \mathrm{softmax}([\mathbf{w}^{C}_{j}, w^{S}_{j}])
\end{equation}

where $\mathbf{w}_{j} \in \mathbb{R}^{N_{C}+1}$ stands for the weights of both task-common and task-specific experts after normalization. In particular, the first $N_{C}$ elements denote weights of task-common experts while the last one represents weight of task-specific one. Besides, it is worthy noting that the linear transformation matrices (i.e., $\mathbf{W}^{C}_{T}$ and $\mathbf{W}^{S}_{T}$) can be various for different layers and it can increase the ability of representation and on the other hand, it is likely to cause over-parameterization.

Regarding a conventional gate function as in CGC \cite{Tang2020progressive}, the input vector $x_{j}$ serves as the variable to the gate function and consequently, the output of the gate function (i.e., contribution weights of experts) varies among samples. Be contrast to that design, the proposed task-motivated one is fed by task IDs instead of the input vector $x_{j}$, which offers an admirable feature of being invariant to the input sample. Also thanks to another characteristic of linear combination among two types of experts, the corresponding parameters learned for a certain task $\mathcal{T}_{j}$ can be expressed as a linear process:

\begin{equation} \label{weight}
\begin{split}
  \mathbf{W}_{j} &= \mathbf{W}_{0} + \frac{\alpha}{r} \cdot \Delta \mathbf{W}_{j} \\
  &= \mathbf{W}_{0} + \frac{\alpha}{r} \cdot [w^{S}_{j} \cdot E^{S}_{j} + \sum^{N_{C}}_{i=1} w^{C}_{ji} \cdot E^{C}_{i}] \\
  &= \mathbf{W}_{0} + \frac{\alpha}{r} \cdot [w_{j} \cdot \mathbf{B}^{S}_{j} \mathbf{A}^{S}_{j} + \sum^{N_{C}}_{i=1} w^{C}_{ji} \cdot \mathbf{B}^{C}_{i} \mathbf{A}^{C}_{i}]
\end{split}
\end{equation}

According to Equation (\ref{weight}), the final parameters of a CGC-LoRA layer of a given task $\mathcal{T}_{j}$ (i.e., $\mathbf{W}_{j}$) is only determined by task ID $j$, which signifies that a unique set of parameters can be reclaimed for all samples from the task $\mathcal{T}_{j}$. In contrast, if the input vector $x_{j}$ also makes an effect on the gate function, the weight matrices $\mathbf{W}_{j}$ would alter across samples from even the same task. Such a tiny distinction definitely leads to high inference latency since distinct samples, even those from the same task $\mathcal{T}_{j}$, have their own unique parameters $\mathbf{W}_{j}$ and parameters need to be calculated repetitively. Considering our task-motivated gate function, the parameters of a given task $\mathcal{T}_{j}$ stay static for samples belonging to that task and during the inference process, all samples from a certain task can be conducted as a batch. Following such an operation, the latency caused by reasoning the parameters can be significantly decreased by $\frac{N_{T}}{N_{sample}}$ in theory. Here, $N_{sample}$ denotes the total numbers of inference samples.

\subsection{Training and Inference} \label{training}
In this section, we elaborate the details of the training and inference process of the proposed CGC-LoRA structure. Algorithm \ref{algorithm} summarizes an entire pipeline and anyone can replicate the training and inference process by following those steps.

\textbf{Training}. Firstly, the pre-trained central LLM and the layers that demand fine-tuning using CGC-LoRA algotithm should be defined (line 1). Next, multiple hyper-parameters, such as rank value $r$, scale value $\alpha$, the numbers of task-common experts, etc., should be determined in advance (line 2-5). During the fine-tuning procedure, full parameters of the pre-trained LLM stay frozen (line 6) and as suggested in \cite{Liu2023moelora}, a batch of training samples are randomly sampled without replacement from various tasks iteratively (line 7). Moreover, we behave the forward process and calculate the loss function based on batches of training samples (line 8-9). Finally, the parameters of CGC-LoRA module and task-motivated gate function (i.e., the parameters of CGC-LoRA $\{ \mathbf{A}^{S}_{i}, \mathbf{B}^{S}_{i} \}^{N_{S}}_{i=1}$ and $\{ \mathbf{A}^{C}_{i}, \mathbf{B}^{C}_{i} \}^{N_{C}}_{i=1}$ and the parameters of task motivated gate function $\mathbf{E}, \mathbf{W}^{C}_{T}$, and $\mathbf{W}^{S}_{T}$) are updated according to the backpropagation mechanism.

\textbf{Inference}. After being fine-tuned, the updated parameters corresponding to each task $\mathcal{T}_{j}$ can be retrieved by following Equation (\ref{weight}). Once the fine-tuned parameter matrices of each task are recovered by steps described in line 12-15, we can apply the corresponding parameters for inference, given a specific task $\mathcal{T}_{j}$.

\begin{algorithm}
  \caption{The training and inference process of CGC-LoRA}\label{algorithm}
  \begin{algorithmic}[1]
  \State Specify the pre-trained central LLM and the layers that demand CGC-LoRA fine-tuning.
  \State Specify the rank value $r$ and scale value $\alpha$.
  \State Specify the numbers of task-common experts $N_{C}$ of CGC-LoRA. (Note: The numbers of task-specific experts $N_{S}$ is set to $N_{S} = N_{T}$)
  \State Specify the rank value $r_{i}$ of each expert. Note: (i) By default, $r_{i} = \frac{r}{N}$. (ii) $\sum^{N}_{i=1}r_{i} = r$.
  \State Specify if the transformation matrices $\mathbf{W^{C}_{T}}$ and $\mathbf{W^{S}_{T}}$ stay the same for different layers: Yes or No.
  \Require
  \State Freeze all parameters in the pre-trained LLM, \textit{e.g.,} $\mathbf{W}_{q}$, $\mathbf{W}_{k}$, $\mathbf{W}_{v}$, and $\mathbf{W}$.
  \For{a batch of samples $\mathcal{S}$ in $\mathcal{D}$}
  \State Execute forward process for LLM guided by CGC-LoRA scheme following Equation (\ref{cgc_forward}).
  \State Calculate the loss function by Equation (\ref{loss}).
  \State Update the parameters of CGC-LoRA $\{ \mathbf{A}^{S}_{i}, \mathbf{B}^{S}_{i} \}^{N_{S}}_{i=1}$ and $\{ \mathbf{A}^{C}_{i}, \mathbf{B}^{C}_{i} \}^{N_{C}}_{i=1}$ and the parameters of task-motivated gate function $\mathbf{E}, \mathbf{W}^{C}_{T}$, and $\mathbf{W}^{S}_{T}$
  \EndFor
  \Ensure
  \For{$\mathcal{T}_{j}$ in $\mathbb{T}$}
  \State Calculate the contribution weights $\mathbf{w}^{C}_{j}$ for task-common experts and $w^{S}_{j}$ for task-common experts following Equation (\ref{wcj}) and Equation (\ref{wsj}), respectively and compute the normalized weights $\mathbf{w}_{j}$ by Equation (\ref{softmax}).
  \State Retrieve the parameters of the fine-tuned LLM with CGC-LoRA scheme by Equation (\ref{weight}) for each task $j$.
  \EndFor
  \State Given a specific task $\mathcal{T}_{j}$, apply the corresponding parameters of the fine-tuned LLM for inference.
  \end{algorithmic}
\end{algorithm}

\section{Experiments}
To comprehensively prove the effectiveness of the proposed CGC-LoRA structure, we conduct well-designed experiments on two public multi-task datasets: (i) Prompt Chinese Biomedical Language Understanding Evaluation (PromptCBLUE) dataset \cite{Zan2021building, Zhu2023promptcblue}. (ii) Firefly dataset \cite{Firefly}. In Section \ref{dataset}, an elaborate introduction to PromptCBLUE and Firefly Dataset is presented and a detailed description of baselines is given in Section \ref{baselines}. Next, the implementation details, such as the settings of software, hardware and hyper-parameters, are illustrated in Section \ref{implementation}. Moreover, Section \ref{evaluation} and Section \ref{performance} amplify the evaluation indexes and the corresponding results of different methods on each task of those two public datasets, respectively. In addition, the conclusion of an ablation study on each component of CGC-LoRA is made in Section \ref{ablation_study} while the effects of hyper-parameter selection are debated in Section \ref{hyperparameter_analysis}.

\subsection{Dataset} \label{dataset}
To the best of our knowledge, Prompt Chinese Biomedical Language Understanding Evaluation dataset \cite{Zan2021building, Zhu2023promptcblue} and Firefly dataset \cite{Firefly} are two most thorough multi-task datasets in Chinese, both of which include numerous tasks of noticeable diversity, such as entity recognition, text generation, text classification, etc.

\textit{PromptCBLUE dataset:} It is derived from Chinese Biomedical Languange Understanding Evaluation (CBLUE) dataset by transforming all samples into a pure text format as described in Section \ref{input}. PromptCLUE is released by East China Normal University on Tianchi Competition Platform \cite{Tianchi} and totally, it contains 16 medicine-related tasks, such as medical named entity recognition, diagnosis report generation, etc. In order to compare with MOE-LoRA \cite{Liu2023moelora} fairly and rationally, we reuse exactly the same training/validation/test datasets as MOE-LoRA and also follow its pre-processing procedure. The detailed statistics of datasets corresponding to 8 selected tasks are summarized in Table \ref{tab:cblue}.

\begin{table}
\setlength{\abovecaptionskip}{0.2cm}
\setlength{\belowcaptionskip}{0.2cm}
\caption{The brief task description and statistics of 8 selected sub-datasets from PromptCBLUE.}
\label{tab:cblue}
\centering
\renewcommand\arraystretch{1.2}
\scalebox{1}{
\begin{tabular}{ccccc}
\toprule[1.5pt]
Task & Description & \# Train & \# Validation & \# Test\\
\midrule[1pt]
CMeIE & Name Entity Recognition & 2828 & 600 & 600\\
CHIP-CDN & Normalization & 2381 & 600 & 600\\
CHIP-CDEE & Attribute Extraction & 1562 & 600 & 600\\
CHIP-MDCFNPC & Clinic Entity Discovery & 4935 & 600 & 600\\
CHIP-CTC & Medical Text Classification & 3622 & 1100 & 1100\\
KUAKE-QIC & Query Intention & 3279 & 660 & 660\\
IMCS-V2-MRG & Report Generation & 1799 & 600 & 600\\
MedDG & Doctor Dialogue & 4964 & 600 & 600\\
\bottomrule[1.5pt]
\end{tabular}}
\end{table}

\textit{Firefly dataset:} It is a comprehensive, public, multi-task dataset in Chinese, totally including over one million samples \cite{Firefly}. Specifically, it contains 23 general tasks, such as lyric generation, sentiment analysis, natural language inference, etc., from which we randomly select 8 tasks (Text Correction (TC), Summary, Keyword Recognition (KR), Text Matching (TM), Sentiment Analyze (SA), MRC, NER, and NLI) and considering the computing complexity, we fix the number of training, evaluation, and test samples of those 8 tasks at 1800, 300, and 300, respectively.To comprehensively evaluate the proposed framework in various situations, we do not transform samples into pure linguistic text format as PromptCBLUE and instead samples stay in an input-target format as displayed in Table \ref{tab:firefly example}.

\begin{table*}[!ht]
\setlength{\abovecaptionskip}{0.2cm}
\setlength{\belowcaptionskip}{0.2cm}
\caption{Examples from Firefly dataset on four tasks: Natural Language Inference, Name Entity Recognition, Sentiment Analysis, and Couplet.} \label{tab:firefly example}
\centering
\renewcommand\arraystretch{1.5}
\scalebox{0.83}{
\begin{tabular}{c|c|c}
\toprule[1.5pt]
Type & Input & Target \\
\midrule[1pt]
Natural Language Inference & Text:soccer game with multiple males playing. Hypothesis:Some men are playing a sport. & Entailment \\
Named Entity Recognition & Beijing is the capital of China. Please recognize all city entities in the sentence. & Beijing \\
Sentiment Analysis & This book is really exciting. What is the sentiment of this comment, positive or negative? & Positive \\
Couplet & Monk. Please give the second line of a couplet: & Wukong \\
\bottomrule[1.5pt]
\end{tabular}}
\end{table*}

\subsection{Baselines} \label{baselines}
To demonstrate the proposed framework powered by the CGC-LoRA module, we conduct a well-designed experiment to compare it with three types of baselines:

\begin{itemize}
\item \textbf{LLM without fine-tuning:} Regarding the baseline without fine-tuning, ChatGPT \cite{Liu2023, Long2022training} acts as the fundamental pre-trained LLM. In specific, to pilot ChatGPT to generate the output in a desired format, we apply In-Context Learning algorithm \cite{Dong2023a} by providing 3 to 10 randomly sampled examples as the task description.
\item \textbf{LLM with LoRA:} As a well-tested fine-tuning algorithm, LoRA is naturally compatible with LLMs due to its effectiveness and efficiency \cite{Hu2021lora}. In multi-task learning (MTL) cases, two straightforward scenarios can be implemented with LoRA: (i) LoRA \textit{Full}: All samples from disparate tasks are utilized simultaneously to fine-tune a pre-trained central LLM using LoRA. (ii) LoRA \textit{Single}: a LLM with a unique set of LoRA modules is fine-tuned for each task using its corresponding samples.
\item \textbf{LLM with multi-task LoRA:} For this type of baselines, the basic pre-trained LLMs are fine-tuned by two recently released LoRA algorithms having MTL capability, separately. One is LoRAHub \cite{Huang2023lorahub} that fine-tunes a distinct set of LoRA for each task respectively and combines all fine-tuned LLMs using the weight optimization via gradient-free methods. The other one is MOE-LoRA, which integrates mixture-of-expert (MOE) structure \cite{Jacobs1991} into LoRA algorithm.
\end{itemize}

Finally, it is worthy noting that ChatGLM-6B \cite{Du2021glm} is utilized in both \textit{LLM with LoRA} and \textit{LLM with multi-task LoRA} because of its dominance in Chinese text related competitions.

\subsection{Implementation Details} \label{implementation}
The configuration details are presented as follows:

\begin{itemize}
\item \textbf{Softare and Hardware:} Software configuration is PyTorch 1.12.0 and Python 3.6.5 while hardware configuration is Tesla A100 GPU.
\item \textbf{Basic LLMs:} As for approaches with fine-tuning (i.e., \textit{LLM with LoRA}, \textit{LLM with multi-task LoRA}, and proposed CGC-LoRA), ChatGLM-6B \cite{Du2021glm} serves as the basic LLM.
\item \textbf{Input/Output Length:} The maximum length of input and output are set to 1024 and 196, separately.
\item \textbf{Hyper-parameters:} Batch size and maximum training steps are configured to 64 and 8000, respectively. Besides, the $\alpha$ value is fixed at 0.1.
\item \textbf{Trainable Layers:} Regarding all fine-tuned LLM, the trainable layers are limited within the self-attention (i.e., \textit{Query}, \textit{Key}, \textit{Value} head) and linear layers of ChatGLM-6B. More details are shown as Figure \ref{fig:cgc}.
\end{itemize}

\subsection{Evaluation Metrics} \label{evaluation}
Since both PromptCBLUE and Firefly contain a variety of tasks, different metrics are implemented to evaluate the performance of baselines and proposed CGC-LoRA on each task. As for PromptCBLUE dataset, CMeIE, CHIPCDN, CHIP-CDEE and CHIP-MDCFNPC evaluate the performance using Micro-F1 score and Macro-F1 score is calculated for CHIP-CTC and KUAKE-QIC. In addition, Rouge-L \cite{Lin2004automatic} is applied in text generation tasks, such as report generation and doctor dialogue. To assess the overall performance, the average score over all tasks is implemented. As for Firefly dataset, identical evaluation metrics are applied. In particular, Micro-F1 score is used for Keyword Recognition, MRC, and NER while Macro-F1 score is calculated for Text Matching, NLI, and Sentiment Analyze. Also, for text generation tasks (e.g., Text Correction and Summary), Rouge-L serves as the evaluation metric. Similarly, the average score is used to evaluate the overall performance.

\subsection{Overall Performance} \label{performance}
Table \ref{tab:results promptcblue} and Table \ref{tab:results firefly} declare the thorough experimental results of the proposed CGC-LoRA and baselines on PromptCBLUE dataset and Firefly dataset, respectively. According to the average scores across all tasks of two datasets, a substantial conclusion can be made that the proposed CGC-LoRA persistently surpasses all baselines and also achieves the most robust performance across a variety of tasks on two totally different datasets. More specific investigation to the experimental results is presented as follows:

\begin{itemize}
\item \textbf{LLMs without fine-tuning:} As a baseline without any fine-tuning, ChatGPT falls behind other approaches on both PromptCBLUE and Firefly datasets, which, in a large sense, underlines the value of fine-tuning procedure. More specifically, task-specific information can be, to a certain extent, injected into the pre-train LLMs through further fine-tuning, which finally has a positive effect on the comprehensive performance on specific tasks. In spite of the global deficiency, we still can observe that ChatGPT outperforms all other methods on Doctor Dialogue thanks to its excellent competence on dialogue comprehension and generation.
\item \textbf{LLM with LoRA:} For LoRA strategy, we make an attempt to implement two straightforward schemes (i.e., LLM \textit{Full} and LLM \textit{Single}) to adapt it to multi-task learning (MTL). Between them, LLM \textit{Full} is in the lead on nearly all tasks and also the overall performance on PromptCBLUE dataset while a completely opposite achievement is observed on Firefly dataset. We suspect that such a paradoxical conclusion is caused by variable similarity across distinctive tasks on those two datasets and this investigation typically demonstrates the significance and necessity of sharing knowledge across diverse tasks in a reasonable way that is also the inspiration to the proposed CGC-LoRA.
\item \textbf{LLM with multi-task LoRA:} As two types of LoRA strategies that have been proven to be compatible with MTL, LoRAHub and MOE-LoRA serves as two competitive benchmarks. According to the results displayed in Table \ref{tab:results promptcblue} and Table \ref{tab:results firefly}, MOE-LoRA leads the other baselines by a large margin on PromptCBLUE dataset while LoRAHub is ahead of others except for LoRA \textit{Single} on Firefly dataset. Similarly, the contradictory consequence that indicates the sensitivity of algorithms to a wide range of tasks can still be disclosed. What is more important, it further signifies the significance of the network architecture that can promote effective and efficient information sharing across distinct tasks in a robust way.
\item \textbf{CGC-LoRA:} Thanks to the task-specific and task-common experts, CGC-LoRA offers an outstanding and also stable achievement on 16 tasks across two datasets and these 16 tasks cover widespread applications in natural language processing. Specifically, it widely leads the other approaches on 10 out of 16 tasks and also stays in a pioneering position on overall performance, which exposes its effectiveness, efficiency and also robustness.
\end{itemize}

In summary, according to the experimental results displayed in Table \ref{tab:results promptcblue} and Table \ref{tab:results firefly}, the proposed CGC-LoRA demonstrates its preferable achievement over basic LoRA strategies (i.e., LoRA \textit{Full} and LoRA \textit{Single}) and multi-task LoRA schemes (i.e., LoRAHub and MOE-LoRA) on these two public datasets.

\begin{table*}
\setlength{\abovecaptionskip}{0.2cm}
\setlength{\belowcaptionskip}{0.2cm}
\caption{The overall results of three kinds of baselines (i.e., LLM without fine-tuning, LLM with LoRA, and LLM with multi-task LoRA) and CGC-LoRA on PromptCBLUE dataset. The boldface stands for the highest score and the underline represents the best result of the baselines. ``$\star$” marks the statistically significant improvements (i.e., two-sided t-test with \textit{p} < 0.05) over the best baseline.}
\label{tab:results promptcblue}
\centering
\renewcommand\arraystretch{1.2}
\scalebox{0.74}{
\begin{tabular}{c|cccccccc|c}
\toprule[1.5pt]
Model & CMeIE & CHIP-CDN & CHIP-CDEE & CHIP-MDCFNPC & CHIP-CTC & KUAKE-QIC & IMCS-V2-MRG & MedDG & Avg. \\
\midrule[1.0pt]
ChatGPT & 0.3058 & 0.6069 & 0.2838 & 0.5854 & 0.5253 & 0.4851 & 0.3253 & \textbf{0.1361} & 0.4067\\
\midrule[1.0pt]
LoRA \textit{Full} & 0.5089 & 0.8748 & 0.5464 & 0.7780 & \textbf{0.8758} & 0.8615 & 0.3678 & 0.1113 & 0.6155\\
LoRA \textit{Single} & 0.4984 & 0.8882 & 0.5528 & 0.7765 & 0.8694 & 0.8524 & 0.3583 & 0.1143 & 0.6138\\
\midrule[1.0pt]
LoRAHub & 0.4411 & 0.8442 & 0.5041 & 0.7177 & 0.8564 & 0.8502 & 0.3061 & 0.1192 & 0.5799\\
MOE-LoRA & \underline{0.5193} & \underline{0.8928} & \underline{0.5697} & $\mathbf{0.7933}^{\star}$ & 0.8691 & \underline{0.8675} & \underline{0.3681} & 0.1089 & \underline{0.6236}\\
\midrule[1.0pt]
CGC-LoRA & $\mathbf{0.5207}$ & $\mathbf{0.8948}$ & $\mathbf{0.5720}$ & 0.7822 & 0.8509 & $\mathbf{0.8727}$ & $\mathbf{0.3808}^{\star}$ & 0.1184 & $\mathbf{0.6240}$\\
\bottomrule[1.5pt]
\end{tabular}}
\end{table*}

\begin{table*}
\setlength{\abovecaptionskip}{0.2cm}
\setlength{\belowcaptionskip}{0.2cm}
\caption{The overall results of three kinds of baselines (i.e., LLM without fine-tuning, LLM with LoRA, and LLM with multi-task LoRA) and CGC-LoRA on PromptCBLUE dataset. The boldface stands for the highest score and the underline represents the best result of the baselines. ``$\star$” marks the statistically significant improvements (i.e., two-sided t-test with \textit{p} < 0.05) over the best baseline.}
\label{tab:results firefly}
\centering
\renewcommand\arraystretch{1.2}
\scalebox{0.85}{
\begin{tabular}{c|cccccccc|c}
\toprule[1.5pt]
Model & TC & Summary & KR & TM & MRC & NER & NLI & SA & Avg. \\
\midrule[1.0pt]
ChatGPT & 0.8799 & 0.2320 & 0.2275 & 0.4700 & 0.6142 & 0.2379 & 0.5408 & 0.9593 & 0.5202\\
\midrule[1.0pt]
LoRA \textit{Full} & 0.9057 & 0.2605 & 0.2740 & 0.7766 & 0.6584 & 0.5500 & 0.7233 & \underline{0.9866} & 0.6418\\
LoRA \textit{Single} & \underline{0.9174} & $\mathbf{0.2745}$ & $\mathbf{0.2744}$ & $\mathbf{0.7966}$ & 0.7222 & 0.5807 & 0.7027 & \underline{0.9866} & \underline{0.6569}\\
\midrule[1.0pt]
LoRAHub & 0.9168 & 0.2681 & 0.2447 & 0.7733 & \underline{0.7310} & \underline{0.5889} & \underline{0.7433} & 0.9833 & 0.6562\\
MOE-LoRA & 0.9093 & 0.2599 & 0.2471 & 0.7833 & 0.7230 & 0.5683 & 0.7333 & 0.9766 & 0.6501\\
\midrule[1.0pt]
CGC-LoRA & $\mathbf{0.9178}$ & 0.2681 & 0.2720 & 0.7866 & $\mathbf{0.7311}$ & $\mathbf{0.6085}^{\star}$ & $\mathbf{0.7533}^{\star}$ & $\mathbf{0.9900}$ & $\mathbf{0.6659}^{\star}$\\
\bottomrule[1.5pt]
\end{tabular}}
\end{table*}

\begin{table*}
\setlength{\abovecaptionskip}{0.2cm}
\setlength{\belowcaptionskip}{0.2cm}
\caption{The experimental results of ablation study for CGC-LoRA. The boldface stands for the highest score. ``$\star$” marks the statistically significant improvements (i.e., two-sided t-test with \textit{p} < 0.05) over the other approaches.}
\label{ablation}
\centering
\renewcommand\arraystretch{1.2}
\scalebox{0.74}{
\begin{tabular}{c|cccccccc|c}
\toprule[1.5pt]
Model & CMeIE & CHIP-CDN & CHIP-CDEE & CHIP-MDCFNPC & CHIP-CTC & KUAKE-QIC & IMCS-V2-MRG & MedDG & Avg. \\
\midrule[1.0pt]
w/o CGC & 0.5089 & 0.8748 & 0.5464 & 0.7780 & 0.8758 & 0.8615 & 0.3678 & 0.1113 & 0.6155\\
w/o gate & 0.5015 & 0.8840 & 0.5378 & 0.7789 & \textbf{0.8818} & 0.8699 & 0.3709 & 0.1140 & 0.6174\\
w multiple gates & \textbf{0.5246} & 0.8874 & 0.5523 & 0.7818 & 0.8300 & 0.8716 & 0.3645 & \textbf{0.1205} & 0.6166\\
\midrule[1.0pt]
CGC-LoRA & 0.5207 & $\mathbf{0.8948}^{\star}$ & $\mathbf{0.5720}^{\star}$ & \textbf{0.7822} & 0.8509 & $\mathbf{0.8727}$ & $\mathbf{0.3808}^{\star}$ & 0.1184 & $\mathbf{0.6240}^{\star}$\\
\bottomrule[1.5pt]
\end{tabular}}
\end{table*}

\subsection{Ablation Study} \label{ablation_study}
To extensively investigate the impact of each component in proposed CGC-LoRA structure, we conduct a comprehensive ablation study on PromptCBLUE dataset and the results are exhibited in Table \ref{ablation}.

\begin{itemize}
\item \textbf{w/o CGC:} As a primary module of CGC-LoRA structure, it plays a compelling role on capturing and processing both task-common and task-specific knowledge. CGC-LoRA will naturally reverts to LoRA \textit{Full} when excluding the CGC architecture. Comparing with CGC-LoRA, \textit{w/o CGC} demonstrates an inferior performance on both task-specific evaluation and overall assessment, highlighting that CGC module critically contributes to the outstanding achievement of proposed CGC-LoRA.
\item \textbf{w/o gate:} Task motivated gate function is responsible for calculating the contribution of task-common and task-specific experts to each task only based on task IDs. For the \textit{w/o gate} variant, uniform expert weights, bypassing the gate function, are employed instead of expert-specific ones. As shown in Table \ref{ablation}, CGC-LoRA exceeds the \textit{w/o gate} variant on 7 out of 8 tasks, which proves the validness of task motivated gate module.
\item \textbf{w multiple gates:} As mentioned in Section \ref{task}, motivated gate function can be unique for each CGC-LoRA layer and it is denoted as the \textit{w multiple gates} variant. From Table \ref{ablation}, we can inspect that CGC-LoRA with multiple gates cannot attain comparable results to the one with single gate on different tasks, in a large sense, due to its over-parameterization and also considering a higher count of trainable parameters brought by multiple gate functions, the proposed CGC-LoRA is designed as a single gate setup.
\end{itemize}

The ablation study demonstrate the effectiveness and necessity of fundamental modules (i.e., CGC and task motivated gate function) of the proposed CGC-LoRA structure, as well as the significance of specific optimization pattern.

\subsection{Hyper-parameter Analysis} \label{hyperparameter_analysis}
To explain the primary hyper-parameter selection in our study, we investigate the influence of the task-common expert number $N_{C}$ while the impact of the task-specific expert number is not considered since it is usually equal to the counts of tasks in MTL. Besides, we also explore the effect of rank of each expert using both task specific evaluation metrics and the average evaluation score on PromptCBLUE dataset. As illustrated in Table \ref{tab:hyperparameter1}, we set the number of task-specific expert and the rank of each expert to 8 and 2, separately. As for the impact of numbers of the task-common expert on the average score, the curve is in a parabola shape and the peak occurs at numbers equal to 8 and this result clarifies that adequate task-common experts are prerequisite to completely capture and efficiently process the task-common knowledge across different tasks. In addition, when fixing the counts of both task-specific and task-common expert at 8, we probe the influence of rank of each expert. Similarly, the curve is also in a parabola shape the the optimal overall achievement happens at rank equal to 2 and detailed results are displayed in Table \ref{tab:hyperparameter2}.

\begin{table*}
\setlength{\abovecaptionskip}{0.2cm}
\setlength{\belowcaptionskip}{0.2cm}
\caption{The experimental results of Hyper-parameter analysis for CGC-LoRA on PromptCBLUE dataset. The number of task-specific experts is fixed at 8, which is equal to the counts of tasks and the rank of each expert is set to 2. The boldface stands for the highest score. ``$\star$” marks the statistically significant improvements (i.e., two-sided t-test with \textit{p} < 0.05) over the other hyper-parameter settings.}
\label{tab:hyperparameter1}
\centering
\renewcommand\arraystretch{1.2}
\scalebox{0.74}{
\begin{tabular}{c|cccccccc|c}
\toprule[1.5pt]
\makecell{\# task-common \\ experts} & CMeIE & CHIP-CDN & CHIP-CDEE & CHIP-MDCFNPC & CHIP-CTC & KUAKE-QIC & IMCS-V2-MRG & MedDG & Avg. \\
\midrule[1.0pt]
2 & 0.4958 & 0.8751 & 0.5532 & 0.7775 & 0.8400 & 0.8564 & 0.3608 & 0.1167 & 0.6094\\
4 & 0.5137 & 0.8922	& 0.5599 & 0.7776 & 0.8309 & 0.8694	& 0.3593 & 0.1159 & 0.6149\\
8 & $\mathbf{0.5206}^{\star}$ & \textbf{0.8947}	& $\mathbf{0.5720}^{\star}$ & \textbf{0.7822} & 0.8509 & 0.8727	& $\mathbf{0.3807}^{\star}$ & \textbf{0.1184} & $\mathbf{0.6240}^{\star}$\\
16 & 0.4984	& 0.8882 & 0.5528 & 0.7765 & $\mathbf{0.8694}^{\star}$ & \textbf{0.8761} & 0.3583 & 0.1143 & 0.6167\\
\bottomrule[1.5pt]
\end{tabular}}
\end{table*}

\begin{table*}
\setlength{\abovecaptionskip}{0.2cm}
\setlength{\belowcaptionskip}{0.2cm}
\caption{The experimental results of Hyper-parameter analysis for CGC-LoRA on PromptCBLUE dataset. The number of task-specific and task-common experts are both fixed at 8 and the rank of each expert stay the same. The boldface stands for the highest score. ``$\star$” marks the statistically significant improvements (i.e., two-sided t-test with \textit{p} < 0.05) over the other hyper-parameter settings.}
\label{tab:hyperparameter2}
\centering
\renewcommand\arraystretch{1.2}
\scalebox{0.8}{
\begin{tabular}{c|cccccccc|c}
\toprule[1.5pt]
rank & CMeIE & CHIP-CDN & CHIP-CDEE & CHIP-MDCFNPC & CHIP-CTC & KUAKE-QIC & IMCS-V2-MRG & MedDG & Avg. \\
\midrule[1.0pt]
1 & 0.5136 & 0.8886 & 0.5625 & \textbf{0.7878} & 0.8463 & 0.7960 & 0.3609 & 0.1147 & 0.6088\\
2 & $\mathbf{0.5206}^{\star}$ & \textbf{0.8947} & 0.5720 & 0.7822 &	\textbf{0.8509} & \textbf{0.8727} & $\mathbf{0.3807}^{\star}$ & \textbf{0.1184} & \textbf{0.6240}\\
4 & 0.5054 & 0.8826	& \textbf{0.5801} & 0.7772 & 0.8481 & 0.8669 & 0.3708 & 0.1146 &	0.6182\\
\bottomrule[1.5pt]
\end{tabular}}
\end{table*}

\section{Related Works}
In this section, we will review two research areas that are directly related to our work: (i) Multi-task learning (MTL) and (ii) Parameter efficient fine-tuning (PEFT). In a large sense, the proposed CGC-LoRA structure intents to unify these two strategies in an innovative way.

\subsection{Multi-Task Learning (MTL)}
Hard parameter sharing \cite{Caruana1997} gives the first and most basic attempt to achieve MTL framework while the unfavorable information transfer caused by straightforward parameters sharing across tasks frequently appears. To solve such a negative transfer issue, cross-stitch network \cite{Misra2016} and sluice network \cite{Ruder122017} are proposed successively, both of which consist of weights of linear combinations and aim to merge knowledge from distinct tasks discriminatorily. Unfortunately, the seesawing problem emerges since knowledge from various tasks is coarsely blended with static weights for all samples. To further enhance the learning efficiency in multi-task cases, several studies with the gate structure and attention mechanism are introduced and such a design intends to make message fusion from individual tasks more flexible and effective. For example, Mixture-of-Experts (MOE) model first presents an innovative architecture that task-common experts are shared at the bottom layer while information from these experts is fused through a gate function at the top layer \cite{Jacobs1991}. Next, Multi-Gate Mixture-of-Experts (MMOE) network extends the single-gate structure in MOE to the multiple-gate one, which can assign personalized fusing weights to different tasks \cite{Ma2018modeling}. Furthermore, Progressive Layered Extraction (PLE) architecture incorporates both task-common and task-specific experts into the model. More specifically, in order to alleviate task conflict and seesawing problems caused by complicated task correlations, parameters of task-common and task-specific experts are isolated in a explicit way, by which can make the convergence during optimization procedure more valid in practice. In this work, we aim to broaden the application scenarios of these impressive schemes to fine-tuning process of large language models (LLMs) to solve similar issues (e.g., the seesawing issue and the task conflict problem).

\subsection{Parameter Efficient Fine-tuning}
With a rapid development of large language models (LLMs), a trend of enlarging the counts of parameters to further ameliorate LLMs becomes unambiguous gradually and as a result, adapting the pre-trained LLMs to tasks in a brand-new field through a full-parameter fine-tuning procedure turns to be prohibitive and impractical. Parameter efficient fine-tuning (PEFT), as an alternative algorithm, aims to degrade the computing cost level by updating partial parameters instead of entire ones. As in \cite{Wang2023codet5, Tinn2023}, a straightforward idea is to only update parameters of selected layers (e.g., the output embedding layer) and meanwhile freeze parameters of the other layers. Alternatively, we can also append one or multiple additional layers to the base model and only parameters of these additional layers are trainable. As a modified version of the latter one, Adapter Tuning \cite{Houlsby2019} incorporates a lightweight adapter network with only a few trainable parameters into LLMs and it demonstrates competitive performance to fine-tuning the top layers. Considering Prefix-tuning \cite{Li2021prefix} and P-tuning \cite{Liu2022ptuning}, one or several task-specific virtual token, serving as an additional, trainable, continuous prompt or embedding, is appended to the original input and compared with the discrete prompt, the continuous one is more feasible and is not restricted to discrete real tokens \cite{Liu2023pretrain}. To further break the limit of sequence length of LLMs, LoRA propose another train of thought, which incorporates a pair of trainable low-rank matrices into each dense layer and it also demonstrates comparable performance to full-parameter fine-tuning \cite{Hu2021lora}. Two outstanding advantages of LoRA are: (i) Low computing cost during fine-tuning. (ii) High efficiency during inference. However, LoRA can only learn paired integral matrices for all tasks that is not an effective and efficient way to capture the task-common and task-specific knowledge across distinct tasks in MTL. In this work, we proposed a framework to implement $1+N$ multi-task fine-tuning pattern in LLMs powered by CGC-LoRA network and it takes advantage of both MTL (e.g., MMOE and PLE) and PEFT (e.g., P-tunig and LoRA) and reveals its parameter efficient fine-tuning capability in MTL cases.

\section{Conclusion}
In this work, we primarily take advantage of both parameter efficient fine-tuning (PEFT) approaches (i.e., LoRA algorithm) and multi-task learning (MTL) strategies (i.e., CGC network). In specific, we propose a general framework that implements $1+N$ multi-task fine-tuning pattern in LLMs. Firstly, a variety of tasks are grouped into $N$ clusters based on professional prior information or Inter-Task Affinity. Next, we release a novel multi-task PEFT method, named by CGC-LoRA, and a pre-trained central LLM is fine-tuned with $N$ one-to-one sets of CGC-LoRA modules on those clusters of tasks. Furthermore, a CGC-LoRA module consists of task-common and task-specific experts that can extract and process the common and specific knowledge across various tasks. Moreover, we also design a task-motivated gate function to decide the contributions of two types of experts to a given task. Since the gate function is only fed by task IDs, parameters of all experts can be represented in a unified way as illustrated in Section \ref{task}. Furthermore, inheriting from PEFT methods, the proposed CGC-LoRA offers an efficient way, in which a pre-trained LLM can be fine-tuned using a small counts of additional trainable parameters. To demonstrate the proposed framework powered by CGC-LoRA module, we conduct comprehensive experiments on two public datasets: (i) PromptCBLUE dataset: It is a medical specific dataset in Chinese; (ii) Firefly dataset: It is a general dataset including a variety of tasks also in Chinese. The experimental results definitely prove the effectiveness and efficiency of CGC-LoRA module. In the future, we plan to further explore the feasibility of implementing the proposed framework in MTL case of even higher diversity.

\bibliographystyle{unsrt}  
\bibliography{references}  






\end{CJK}
\end{document}